\documentclass[11pt]{article}

\usepackage[preprint]{acl}

\usepackage{times}
\usepackage{latexsym}
\usepackage[T1]{fontenc}

\usepackage[utf8]{inputenc}

\usepackage{microtype}

\usepackage{inconsolata}

\usepackage{graphicx}
\usepackage{amsmath}
\usepackage{dsfont}
\usepackage{booktabs}
\usepackage{multirow}
\usepackage{shortcuts}
\usepackage{float}
\usepackage{array}
\usepackage{listings}
\lstdefinestyle{prompt}{basicstyle=\ttfamily\scriptsize, columns=fullflexible, keepspaces=true, breaklines=true, breakindent=0pt, frame=single, framesep=3pt, aboveskip=2pt, belowskip=6pt}

\usepackage{enumitem}
\setlist{nosep, leftmargin=*}

\makeatletter
\g@addto@macro\normalsize{
  \setlength\abovedisplayskip{2pt plus 2pt minus 0pt}%
  \setlength\belowdisplayskip{2pt plus 2pt minus 0pt}%
  \setlength\abovedisplayshortskip{0pt plus 2pt}%
  \setlength\belowdisplayshortskip{2pt plus 2pt}}
\makeatother

\title{Can This LLM Annotate My Data?\\ Estimating Zero-Shot LLM Performance}

\author{
    Tong Lin\textsuperscript{1},  
    Grant Schoenebeck\textsuperscript{1}, 
    and Sabina Tomkins\textsuperscript{1} \\
    \textsuperscript{1}School of Information, University of Michigan \\
    \texttt{\{tonglin, schoeneb, stomkins\}@umich.edu}
}

\begin{document}
\maketitle
\begin{abstract}
Large language models (LLMs) are increasingly used in annotation tasks with varying success.
While several studies have demonstrated the opportunities and shortcomings of LLMs for annotation, a critical question remains unanswered: is it possible to predict how well they will perform on a new dataset?
To answer this and related questions, we construct two task families which induce 461 text classification datasets. We find that it is possible to predict LLM performance - to a certain degree - and that contrary to other work which highlights the usefulness of LLM verbalizations of self-confidence, internal log-probabilities are the most predictive signal. 
\end{abstract}

\section{Introduction}

Large language models (LLMs) are increasingly used as annotators to produce labeled data at scale for natural language processing (NLP) and the social sciences \citep{gilardi2023chatgpt, ziems-etal-2024-large}. Yet in many  settings, a practical question arises before annotation begins: given a dataset, how well will an LLM label the data under a reasonable prompting setup? Answering this question early matters as annotation pipelines often require substantial investment such as collecting gold labels or human spot-checking, and the costs of misjudging annotation quality can propagate into downstream modeling and scientific conclusions. 

One way to answer this question is to look at signals the LLM already produces while it labels. A growing body of work studies such signals for flagging individual incorrect predictions. Answering this instance-level question is useful for selective prediction \citep{chen2023adaptation}, or for routing individual hard cases to human review \citep{wang2024human}. But that alone does not answer the question a practitioner facing a new, unlabeled dataset needs answered: not ``which predictions are wrong,'' but ``is this dataset labeled reliably enough to use?''
A few studies estimate a model's accuracy on a new dataset \citep{garg2022leveraging, fu-etal-2023-estimating, wang2024plug, chen2025evaluating}, but each focuses on a single kind of signal. It remains unclear which signals best support such a decision and how many instances are needed.
Thus we ask:
\begin{itemize}
\item Can LLM-derived signals support actionable decisions about LLM labeling?
\item Which signals best support decisions about whether LLM labeling is reliable for a dataset?
\item How does the number of sampled instances influence these decisions?
\end{itemize}

To answer these questions we create two task families which allow us to generate 461 text classification datasets with metadata-derived ground truth: (1) headline topic differentiation and (2) movie script genre classification. 
We then vary the number of instances sampled from each dataset to understand how much data is needed to predict whether an LLM can label it with at least a certain accuracy. We contribute the following: 

\begin{itemize}
    \item We find that we can predict whether an LLM will have classification accuracy of at least 0.9 better than trivial
baselines for all three LLMs we test (with gains of 10-25\% over baselines).
    \item We consistently see that internal probability signals perform better than verbalized signals generated by the LLM.
    \item We find that there are diminishing returns after $\sim$100 sampled instances, that is, performance can often be predicted well with only $\sim$100 samples. 
\end{itemize}

\begin{figure*}[h!]
\centering

\resizebox{1\textwidth}{!}{
    \includegraphics{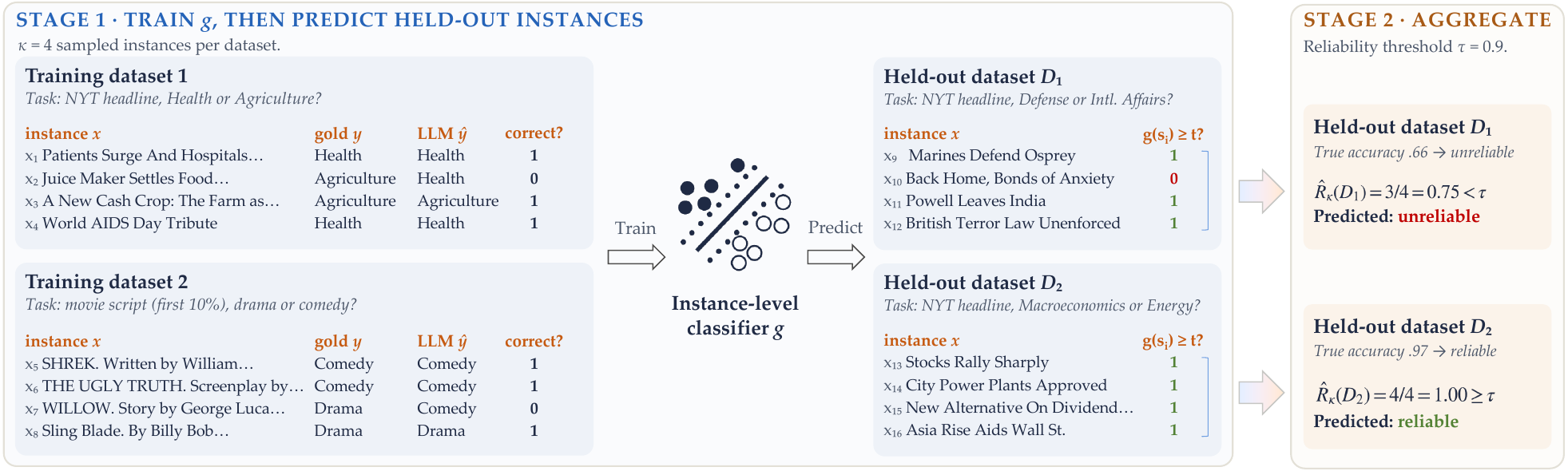}
}

\caption{Two-stage estimator: from instance-level correctness predictions to a dataset-level reliability decision.}
\label{fig:flow}
\end{figure*}

\paragraph{Related Work}
LLM annotation quality varies across datasets and tasks \citep{pangakis2023automated}, and annotation errors can distort downstream conclusions \citep{baumann2025large}. The common practice is to compare LLM labels against human annotations, either to validate the LLM \citep{pangakis2023automated, calderon-etal-2025-alternative} or to debias the analyses built on LLM labels \citep{egami2023using}. Another line of work estimates the reliability of individual predictions from signals the LLM produces while labeling: token probabilities \citep{kadavath2022language}, consistency across repeated runs \citep{manakul-etal-2023-selfcheckgpt, kuhn2023semantic} and verbalized confidence \citep{lin2022teaching, tian2023just, xiong2024can}. However, insights on which family of signals is most informative about LLM annotation performance are mixed: for example, \citet{tian2023just} find verbalized confidence better calibrated than token probabilities, 
whereas \citet{xiong2024can} find it overconfident and concentrated on a few values \citep{merdjanovska2026evaluation}.
None of this work compares different signals at the dataset level. 



Closer to our work, \citet{chen2025evaluating} show that a score based on output consistency tracks annotation accuracy across datasets and  \citet{wang2024plug} use likelihood and perplexity features for the same purpose. \citet{fu-etal-2023-estimating} train a dataset-level meta-model that maps the distribution of probability scores in a dataset to its accuracy. ATC \citep{garg2022leveraging} instead works at the instance level by estimating a new dataset's accuracy as the share of its instances whose confidence exceeds a cutoff learned on labeled data. 
We advance this work with a comprehensive comparison of the signals which predict LLM annotation performance.


\section{Our Approach}
Let $D = \{x_1, \dots, x_n\}$ be a dataset of instances, each requiring a label from a task-specific label set $\mathcal{Y}$. An LLM annotator $\ell$ is prompted to label each $x_i$; querying $\ell$ multiple times per instance and aggregating the responses (by majority vote) yields a predicted label $\hat{y}_i$.

Alongside $\hat{y}_i$, this process yields a signal vector $s_i$ for each instance. Crucially, $s_i$ requires neither a gold label nor any inspection of $x_i$'s content beyond what $\ell$ already used to produce $\hat{y}_i$.

We define a dataset's accuracy as how well $\ell$'s aggregated predictions match the ground truth:

\[
R(D) = \frac{1}{n}\sum_{i=1}^n \mathds{1}[\hat{y}_i = y_i].
\]

In our experiments every dataset has gold labels $y_i$, so $R(D)$ is computable. We define a dataset's reliability as the binary indicator
\[
r(D) = \mathds{1}[R(D) \geq \tau],
\]
for a threshold $\tau$ reflecting a practitioner's tolerance for label noise: $D$ is reliable if $\ell$'s labels on it are accurate enough to use, and unreliable otherwise.

We study a collection of $M$ such datasets $\mathcal{D} = \{D^{(1)}, \dots, D^{(M)}\}$, each independently labeled by $\ell$ following the procedure above, with its own reliability label $r(D^{(m)})$. 
Our goal is to predict $r(D)$ for a held-out dataset $D \in \mathcal{D}$ from the signals of $\kappa$ instances $S_\kappa(D)$ sampled from it, with no access to their gold labels or to the content of $D$. We do this in two stages (\figref{fig:flow}). First, an instance-level classifier $g$ learns whether the LLM labels an instance correctly: given the signals $s_i$ of an instance, it outputs the probability $g(s_i)$ that $\hat{y}_i = y_i$. It is trained on $\kappa$ sampled instances from each training dataset of the fold. Second, we turn these instance-level predictions into a dataset-level decision. Following ATC \citep{garg2022leveraging}, we estimate the accuracy of $D$ as the share of its sampled instances that $g$ predicts to be correct,
\[
\hat{R}_\kappa(D) = \frac{1}{\kappa}\sum_{x_i \in S_\kappa(D)} \mathds{1}[g(s_i) \geq t],
\]
and predict $D$ to be reliable if $\hat{R}_\kappa(D) \geq \tau$. The cutoff $t$ is set on the training instances so that the share of them with $g(s_i) \geq t$ equals their accuracy. We report results separately for $g$ restricted to different subsets of signals.


\subsection{Predictive Signals (Features)}

For each instance $x_i$ in each $D^{(m)}$, $\ell$ is queried $k$ times, yielding a label and its associated signals for each run $j = 1, \dots, k$.

\paragraph{Internal Probability.} Each run $j$ exposes the log-probability $\ell$ assigned to the tokens of its  label $\hat{y}_i^{(j)}$. We record two quantities, averaged across the $k$ runs and negated so higher values indicate greater uncertainty: the log-probability of the label's first token 
(writing $w^{(j)}_{i,1}$ for the first token of $\hat{y}_i^{(j)}$),
\[
\uncertainty_{\mathrm{first}}(x_i) = -\frac{1}{k}\sum_{j=1}^k \log p_j\big(w^{(j)}_{i,1}\big),
\]
and the average log-probability across all tokens of the label,
\[
\uncertainty_{\text{avg}}(x_i) = -\frac{1}{k}\sum_{j=1}^k \overline{\log p_j}\big(\hat{y}_i^{(j)}\big).
\]
Both require access to the log-probability $\ell$ assigns to its own  output tokens,
which open-weight
models and some APIs (here GPT-4o mini) expose.

\paragraph{Label Consistency.} Alongside the internal probability signals, we compute one label consistency signal, self-agreement: the fraction of the $k$ runs whose label matches the majority-vote label $\hat{y}_i$,
\[
\agreement(x_i) = \frac{1}{k}\sum_{j=1}^k \mathds{1}\big[\hat{y}_i^{(j)} = \hat{y}_i\big].
\]

\paragraph{Verbalized Signals.} Beyond the label, $\ell$ is also prompted in the same structured response to self-report a confidence score and a stability score on each run (see Appendix~\ref{app:prompt}). 
We average these across  $k$ runs to obtain two  signals:, $\confidence(x_i)$ and $\stability(x_i)$; stability is $\ell$'s own estimate of how likely it is to assign the same label again.
We retrain $g$ for each $\kappa$, where each dataset contributes $\kappa$ instances in training and testing.

\commentout{
We average these across the $k$ runs to obtain two  signals:

\begin{itemize}
    \item \textbf{Aggregated confidence.} $\confidence_i^{(j)} \in [0,1]$ is $\ell$'s self-reported confidence on run $j$:
    \[
    \confidence(x_i) = \frac{1}{k}\sum_{j=1}^k \confidence_i^{(j)}.
    \]
    \item \textbf{Aggregated stability.} $\stability_i^{(j)} \in [0,1]$ is $\ell$'s self-reported stability on run $j$, indicating how likely it is to assign the same label again:
    \[
    \stability(x_i) = \frac{1}{k}\sum_{j=1}^k \stability_i^{(j)}.
    \]
\end{itemize}
}

\begin{table*}[t]
\centering
\footnotesize
\setlength{\tabcolsep}{2pt}
\resizebox{\textwidth}{!}{%
\begin{tabular}{ll cccccc cccccc}
\toprule
 & & \multicolumn{6}{c}{Accuracy} & \multicolumn{6}{c}{F1} \\
\cmidrule(lr){3-8} \cmidrule(lr){9-14}
\multicolumn{2}{r}{$\kappa$} & 10 & 50 & 90 & 130 & 170 & 210 & 10 & 50 & 90 & 130 & 170 & 210 \\
\splitrule
\multicolumn{2}{l}{\textit{Baseline}} & \multicolumn{6}{c}{\textit{.740}} & \multicolumn{6}{c}{\textit{.850}} \\
\multicolumn{2}{l}{ATC \citep{garg2022leveraging}} & $.717_{.015}$ & $.750_{.016}$ & $.766_{.004}$ & $.774_{.011}$ & $.781_{.012}$ & $.782_{.005}$ & $.815_{.011}$ & $.830_{.012}$ & $.841_{.003}$ & $.847_{.008}$ & $.851_{.008}$ & $.852_{.003}$ \\
\multicolumn{2}{l}{Meta-model \citep{fu-etal-2023-estimating}} & $.666_{.015}$ & $\mathbf{.772_{.016}}$ & $\mathbf{.790_{.017}}$ & $.795_{.015}$ & $\mathbf{.810_{.008}}$ & $.806_{.019}$ & $.760_{.009}$ & $.846_{.013}$ & $.855_{.013}$ & $.861_{.011}$ & $\mathbf{.871_{.005}}$ & $.866_{.013}$ \\
\splitrule
\multirow{2}{*}{Internal probability} & $\mathrm{uncertainty}_{\mathrm{first}}$ & $.726_{.020}$ & $.767_{.019}$ & $.787_{.005}$ & $\mathbf{.808_{.008}}$ & $.808_{.003}$ & $\mathbf{.814_{.006}}$ & $.822_{.015}$ & $.844_{.015}$ & $.856_{.005}$ & $\mathbf{.872_{.006}}$ & $\mathbf{.871_{.002}}$ & $\mathbf{.875_{.004}}$ \\
 & $\mathrm{uncertainty}_{\mathrm{avg}}$ & $.714_{.014}$ & $.762_{.011}$ & $.762_{.011}$ & $.781_{.006}$ & $.787_{.005}$ & $.790_{.002}$ & $.813_{.010}$ & $.841_{.008}$ & $.840_{.008}$ & $.854_{.005}$ & $.857_{.003}$ & $.859_{.000}$ \\
\splitrule
\multirow{1}{*}{Label consistency} & agreement & $\mathbf{.754_{.003}}$ & $.769_{.005}$ & $.769_{.002}$ & $.774_{.006}$ & $.773_{.002}$ & $.773_{.001}$ & $\mathbf{.852_{.002}}$ & $\mathbf{.863_{.002}}$ & $\mathbf{.864_{.001}}$ & $.867_{.003}$ & $.866_{.001}$ & $.866_{.000}$ \\
\splitrule
\multirow{2}{*}{Verbalized} & confidence & $.677_{.003}$ & $.662_{.008}$ & $.648_{.005}$ & $.659_{.008}$ & $.645_{.008}$ & $.638_{.005}$ & $.786_{.003}$ & $.764_{.008}$ & $.748_{.005}$ & $.760_{.008}$ & $.748_{.006}$ & $.742_{.005}$ \\
 & stability & $.673_{.018}$ & $.661_{.010}$ & $.653_{.008}$ & $.658_{.006}$ & $.657_{.007}$ & $.644_{.005}$ & $.782_{.015}$ & $.762_{.010}$ & $.752_{.009}$ & $.758_{.005}$ & $.756_{.005}$ & $.745_{.005}$ \\
\splitrule
\multicolumn{2}{l}{All five signals} & $.719_{.018}$ & $.757_{.007}$ & $.759_{.008}$ & $.782_{.006}$ & $.779_{.004}$ & $.787_{.005}$ & $.821_{.013}$ & $.838_{.005}$ & $.837_{.006}$ & $.853_{.005}$ & $.851_{.002}$ & $.855_{.003}$ \\
\bottomrule
\end{tabular}}
\caption{Accuracy and F1 of the estimator for GPT-4o mini at $\tau = 0.9$, with HGB as instance-level classifier, using each signal on its own or all five signals together, for each number of sampled instances $\kappa$. Values are means over three random seeds with standard errors as subscripts. The baseline predicts the majority class for accuracy and every dataset as reliable for F1. ATC and the meta-model use average-token uncertainty. Bold: best in column.}
\label{tab:results}
\end{table*}

\begin{figure*}[h!]
\centering
\resizebox{\textwidth}{!}{
    \includegraphics{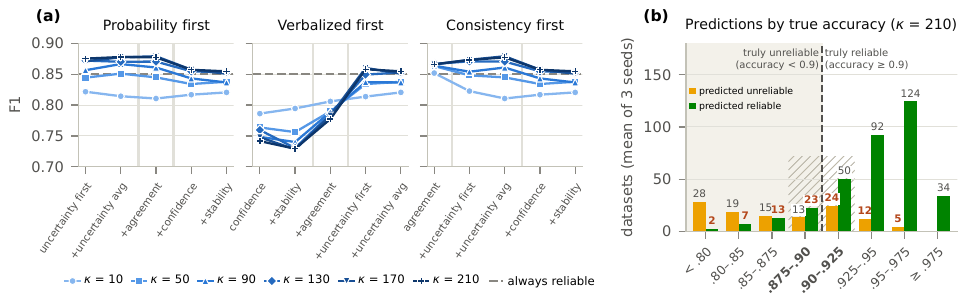}
}
\caption{(a) F1 of the reliability estimator for GPT-4o mini at $\tau = 0.9$ as signals are added one at a time (left to right), in three orders that each start from a different signal family; each line is one number of sampled instances $\kappa$. (b) Number of datasets in each bin of true accuracy that the estimator with the first-token uncertainty alone predicts reliable or unreliable at $\kappa = 210$. }
\label{fig:incremental}
\end{figure*}

\section{Experiments}
We build two  families of text classification tasks whose ground truth comes from structured metadata. Together, these two families give $M = 110 + 351 = 461$ datasets, and let us test whether signal-based reliability estimation holds up under two distinct types of tasks where we can manipulate the difficulty level (e.g. by varying the amount of input data or  topic similarity).

\begin{description}
    \item[Sliding-window genre pairs.]  We use movie scripts paired with each film's genre listing \citep{ismaelmousa2024movies}. For each pair of genres, we take only movies tagged with exactly one of the two genres, excluding movies tagged with both as ambiguous, and ask an LLM which of the two genres the movie belongs to, given all or part of its script. 
    \commentout{We select 11 genre pairs, drawn from genres with at least 50 usable movies each, and generate 10 script-length variants per pair}
    We select 11 genre pairs, each with 247--287 usable movies and at least 38 of each genre (Appendix~\ref{app:data}), and generate 10 script-length variants per pair: the full script (capped at 50{,}000 words) and 9 partial versions covering 10\%--90\% of that capped script, each truncated to the nearest sentence boundary. This yields $11 \times 10 = 110$ datasets. 
    \item[NYT topic pairs.]We use the Comparative Agendas Project's coding of New York Times front-page articles into one of 27 major-topic policy categories \citep{baumgartner2019comparative}. Given only an article's headline, an LLM must decide which of the two topics it belongs to. We construct one dataset for every pair of the 27 topics present in the data, which results in $351$ pairs in total, ranging from topically distant pairs (e.g., Defense vs. Churches and Religion) to closely related ones (e.g., Macroeconomics vs. Banking, Finance, and Domestic Commerce). Each dataset has 300 instances.
\end{description}

\subsection{Evaluation}

In our experiments, $\ell$ is instantiated as three LLMs: GPT-4o mini, accessed via the OpenAI API \citep{openai_api}, and the open-weight Llama-3.1-8B-Instruct \citep{grattafiori2024llama} and Qwen3-8B \citep{yang2025qwen3}, each labeling every instance $k = 5$ times at temperature 1. 
For instance-level classifier $g$, we test SVM, histogram-based gradient-boosted trees (HGB), and XGBoost. Because our datasets reuse texts (e.g., each NYT topic contributes the same 150 articles to all 26 of its pairs, and each genre pair appears at ten script lengths), we use grouped cross-validation so that no test text appears in training (see Appendix~\ref{app:cv}).  

We use $\tau = 0.9$ as the reliability threshold; 74.0\%, 36.2\%, and 54.4\% of the 461
datasets reach it for GPT-4o mini, Llama-3.1-8B,
and Qwen3-8B.
For each $\kappa \in \{10, 50, \dots, 210\}$, we report the metrics over the pooled held-out predictions and average them over three random seeds.

\tabref{tab:results} summarizes the results for GPT-4o mini as the annotator, with HGB as $g$ (see Appendix~\ref{app:full} for all classifiers and LLMs).  Combining all five signals is not the best choice since generally the internal probability or label consistency signals alone do better. The first-token uncertainty is the strongest signal from $\kappa = 130$ on and reaches an F1 of .875 at $\kappa = 210$, with most of the gain reached by $\kappa = 130$. The average-token uncertainty also improves with $\kappa$ but stays below the first-token uncertainty at every $\kappa$. Agreement has the highest F1 for $\kappa \leq 90$, and its F1 levels off at about .865 from $\kappa = 50$. The verbalized signals perform worst, below the baseline at every $\kappa$.

Against the baselines, the margin for GPT-4o mini is modest (at $\kappa = 210$, accuracy .814 for first-token uncertainty against .740 for the majority class and .806 for the meta-model) but larger for Llama-3.1-8B (.802 against .638 and .750) and Qwen3-8B (.729 against .544 and .705). Most errors lie near $\tau$ (\figref{fig:incremental}b): after excluding the datasets within .025 of $\tau$, HGB with uncertainty is the best method for every LLM, and the farther the remaining datasets are from $\tau$, the more often it is correct (Appendix~\ref{app:full}).

\commentout{47 of the 56 datasets with accuracy below .85 are flagged, and 72 of the 86 errors are within .05 of $\tau$.}

We then train the estimator by adding one signal at a time, family by family, in three orders that each start from a different family (\figref{fig:incremental}a). From $\kappa = 130$ onward the three orders show the same pattern. Again the internal probability signals contribute the most: the performance changes little when label consistency is added after them, and rises the most when they are added later. The verbalized signals give the lowest starting performance, which even declines as $\kappa$ grows, and mostly lower the performance when they come later. The label consistency signal gives a modest starting performance at small $\kappa$, and adding the internal probability signals after it helps at larger $\kappa$.

\paragraph{Conclusion}
Our results show that signals from a small sample of instances (less than 150) labeled by the LLM can support the decision of whether to use zero-shot annotation. Among the signals, internal probabilities are the most useful, and the first-token uncertainty alone matches
or outperforms the full signal set. Practitioners should
prefer models that expose token probabilities when possible; if these are not available,
we find that self-agreement is a better predictor 
than verbalized confidence.

\section*{Limitations}
We construct two task families which allow us to derive ground truth from metadata. The advantage of this approach is that we can understand LLM performance in a setting with little label noise; the topic codes
and genre tags are still human-assigned, and we observe that
disagreement about them is plausibly higher
for closely related pairs.
However, future work should re-apply our approach to a larger and more varied collection of datasets:  our estimator is always trained on gold-labeled datasets from the same two task families and prompt format as the test dataset (with disjoint texts), so transfer to an unrelated task is untested. We also do not compare against hand-labeling the $\kappa$ sampled instances, the natural alternative when some annotation budget exists, and our standard errors reflect variation across three seeds, not across datasets. In addition, following prior work, we define reliability through accuracy at a fixed threshold. In scenarios where other aspects of performance matter more, such as performance on a minority class, other metrics like F1 could serve as the reliability measure instead. We test three LLMs and the relative usefulness of the signals already differs among them (see Appendix~\ref{app:full}) and may differ further for other models or decoding settings. Finally, we study only binary text classification; tasks with more labels or free-form outputs, such as spans or summaries, may call for different signals.

\section*{Acknowledgments}
This work is supported by the National Science Foundation under Grant Nos.~2313137 and 2445924. Any opinions, findings, and conclusions or recommendations expressed in this material are those of the author(s) and do not necessarily reflect the views of the National Science Foundation.


\bibliography{custom}

\appendix

\section{Annotation Prompt and Signals}
\label{app:prompt}

Every annotation run sends the system and user prompts in \figref{fig:prompt}. The bracketed fields are filled with the dataset's task description, its two labels with their definitions, and the label names (\tabref{tab:examples} shows two datasets), and with the instance's text, which is a headline for the NYT topic pairs and a script or its truncation for the genre pairs.

GPT-4o mini (\texttt{gpt-4o-mini-2024-07-18}) is queried through the OpenAI API in JSON mode, with the log-probability of every output token returned. Llama-3.1-8B-Instruct and Qwen3-8B run locally with vLLM \citep{kwon2023efficient}, with JSON output enforced during decoding, a top-$p$ of 1, and at most 128 output tokens. Qwen3-8B runs with its thinking mode turned off, and the 1{,}056 instances whose prompts exceed its default context of 40{,}960 tokens are run with YaRN context extension \citep{peng2024yarn}. Each instance is labeled $k = 5$ times at temperature 1.

To read the log-probabilities of the label, we locate the value of the \texttt{"label"} field in the output and take the tokens that overlap it: $\uncertainty_{\mathrm{first}}$ uses the first of these tokens and $\uncertainty_{\text{avg}}$ the mean over all of them. For the open-weight LLMs, these are the log-probabilities of the model's raw output distribution, before temperature scaling. The verbalized signals are the values of the \texttt{"confidence"} and \texttt{"stability"} fields, clipped to $[0, 1]$.

\begin{figure*}[t]
\small
\textbf{System prompt}
\begin{lstlisting}[style=prompt]
You are a careful annotator. Follow the task definition exactly.
Return valid JSON only. Do not include any extra text before or after the JSON.
\end{lstlisting}
\textbf{User prompt}
\begin{lstlisting}[style=prompt]
TASK
You will assign exactly ONE label to the given example.

Task description:
[task description]

Labels (choose exactly one):
[label definitions]

Example to label:
[example text]

Output format (JSON only):
{
  "label": "<one of: [label names]>",
  "confidence": <number between 0 and 1>,
  "stability": <number between 0 and 1>
}

Field definitions:
- "label": your predicted label for this example
- "confidence": how confident you are that the chosen label is correct
- "stability": how likely you think you would give the same label again if asked to annotate this exact same example ten times under the same instructions

Rules:
- Choose exactly one label from the provided list.
- The "label" value must exactly match one of: [label names]
- "confidence" must be a number between 0 and 1.
- "stability" must be a number between 0 and 1.
- Do NOT output explanations.
- Do NOT invent new labels.
- If the example is ambiguous, still choose the single best label and lower confidence and/or stability as appropriate.
- Return a single JSON object with exactly the keys shown above.
\end{lstlisting}
\caption{The system and user prompts of every annotation run. Fields in square brackets are filled in for each dataset and instance (\tabref{tab:examples}); \texttt{[label names]} is the two labels separated by a comma.}
\label{fig:prompt}
\end{figure*}

\section{Datasets}
\label{app:data}

For the NYT topic pairs, each of the 27 topics contributes 150 front-page articles, and the dataset for a pair of topics contains the 150 articles of each. All 351 NYT datasets thus have 300 instances with balanced labels, and the LLM sees only the headline.

For the sliding-window genre pairs, \tabref{tab:genre-pairs} lists the 11 pairs. A pair uses the movies tagged with exactly one of its two genres. We drop movies whose scraped script is shorter than 500 characters, which leaves 247--287 movies per pair, and some pairs are unbalanced (e.g., 38 romance and 249 drama movies). 

\begin{figure*}[t]
\centering
\includegraphics[width=\textwidth]{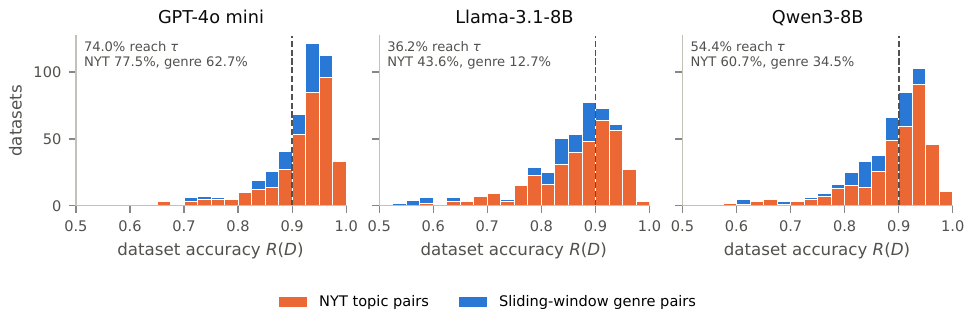}
\caption{Dataset accuracy $R(D)$ of the 461 datasets for each LLM, stacked by task family. The dashed line marks $\tau = 0.9$, and the text gives the share of datasets that reach it, overall and in each family.}
\label{fig:accuracy-distribution}
\end{figure*}

\tabref{tab:examples} shows the task description, the label definitions, and one instance of a dataset from each family. \figref{fig:accuracy-distribution} shows how dataset accuracy is distributed for each LLM. 

\begin{table}[t]
\centering
\footnotesize
\setlength{\tabcolsep}{3pt}
\begin{tabular}{llrr}
\toprule
Genre A & Genre B & $n_A$ & $n_B$ \\
\midrule
mystery \& suspense & drama & 90 & 194 \\
drama & comedy & 144 & 137 \\
drama & action \& adventure & 143 & 130 \\
mystery \& suspense & action \& adventure & 141 & 133 \\
science fiction \& fantasy & action \& adventure & 73 & 189 \\
romance & drama & 38 & 249 \\
romance & comedy & 60 & 220 \\
comedy & action \& adventure & 142 & 134 \\
mystery \& suspense & horror & 207 & 40 \\
science fiction \& fantasy & mystery \& suspense & 104 & 169 \\
science fiction \& fantasy & drama & 105 & 171 \\
\bottomrule
\end{tabular}
\caption{The 11 genre pairs and the number of usable movies of each genre, after dropping movies whose scraped script is shorter than 500 characters. Each pair gives ten datasets, one per script length, with the same movies.}
\label{tab:genre-pairs}
\end{table}

\begin{table*}[t]
\centering
\small
\begin{tabular}{>{\raggedright\arraybackslash}p{0.15\textwidth} >{\raggedright\arraybackslash}p{0.37\textwidth} >{\raggedright\arraybackslash}p{0.37\textwidth}}
\toprule
 & NYT topic pair & Sliding-window genre pair \\
\midrule
Task description & Given this article, which of these topics does it belong to: `Health' or `Agriculture'? & Given this movie script, which genre does the movie belong to: `drama' or `comedy'? \\
\addlinespace
Label definitions & Health: The article belongs to the topic `Health'.\newline Agriculture: The article belongs to the topic `Agriculture'. & drama: The script comes from a movie with genre `drama' and not genre `comedy'.\newline comedy: The script comes from a movie with genre `comedy' and not genre `drama'. \\
\addlinespace
Example text & Patients Surge And Hospitals Hunt for Beds & SHREK Written by William Steig \& Ted Elliott SHREK Once upon a time there was a lovely princess. But she had an enchantment upon her of a fearful sort which could only be broken by love's first kiss. [\dots] (1{,}295 words) \\
\addlinespace
Gold label & Health & comedy \\
\bottomrule
\end{tabular}
\caption{Task description, label definitions, and one instance with its gold label for a dataset of each family, as filled into the prompt of \figref{fig:prompt}. These are the two training datasets of \figref{fig:flow}; the genre dataset uses the first 10\% of each script.}
\label{tab:examples}
\end{table*}

\section{Grouped Cross-Validation}
\label{app:cv}

Our datasets share texts. Each NYT topic contributes the same 150 articles to all 26 of its pairs, the ten lengths of a genre pair contain the same movies, and since a movie can carry several genre tags, different genre pairs can also share movies. Leaving out one dataset at a time would therefore leave test texts in the training data, so we build folds in which no test text appears in training.

For the NYT topic pairs, we shuffle the 27 topics into 9 groups of 3. For each pair of groups $(A, B)$, including $A = B$, the test set is every NYT dataset whose two topics fall in $A$ and $B$, and the training set is every NYT dataset with neither topic in $A$ or $B$, plus all genre datasets. For the genre pairs, each fold tests one genre pair at all ten lengths and trains on all NYT datasets plus the genre pairs that share neither genre with it. Because these pairs can still share movies, each training dataset drops the instances whose text also appears in the test set. This gives $45 + 11 = 56$ folds, and every dataset is tested exactly once.

In each fold, $g$ and its cutoff $t$ are fitted on $\kappa$ instances sampled from each training dataset, and each test dataset is scored on $\kappa$ instances sampled from it. Since every dataset is tested once, we pool the estimates of all 461 datasets and compute accuracy and F1 once per seed. The three seeds (42, 43, and 44) change the topic groups and every sampled set of instances.

\section{Instance-Level Classifiers}
\label{app:impl}

For the instance-level classifier $g$ we test SVM, HGB, and XGBoost. HGB is histogram-based gradient boosting with 200 iterations. XGBoost uses 200 trees of depth at most 6 and a learning rate of 0.1. 

\section{Baselines}
\label{app:baseline}

We compare our estimator with ATC \citep{garg2022leveraging} and the meta-model of \citet{fu-etal-2023-estimating}. Both see the same folds and the same $\kappa$ instances per dataset as our estimator, and use only the gold labels of the training datasets' sampled instances. Both need a confidence score for each instance, and we use $\exp(-\uncertainty_{\text{avg}}(x_i))$ as the closest of our signals to the label probability these methods use. 

\begin{figure*}[t]
\centering
\includegraphics[width=\textwidth]{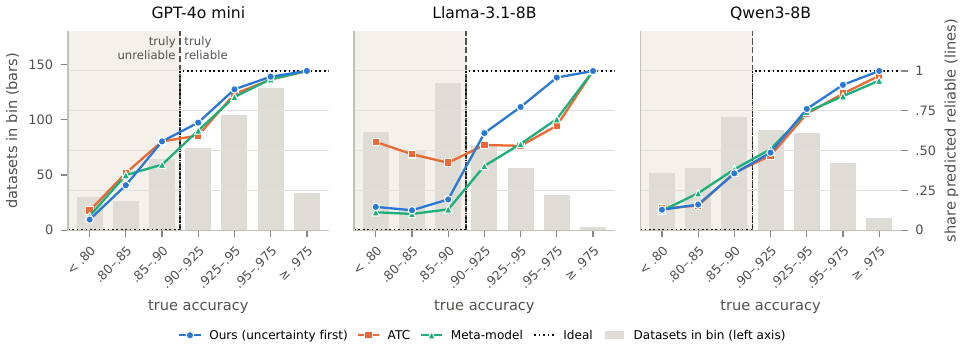}
\caption{Datasets in each bin of true accuracy (bars, left axis) and the share of each bin that each method predicts reliable (lines, right axis) at $\kappa = 210$, averaged over three seeds. Our estimator uses HGB with the first-token uncertainty. Lower lines are better in the shaded bins, which hold the truly unreliable datasets ($R(D) < \tau$), and higher lines are better to the right of the dashed line; the dotted line marks the ideal.}
\label{fig:bins}
\end{figure*}

\figref{fig:bins} shows where the three methods go wrong at $\kappa = 210$. For GPT-4o mini and Qwen3-8B the three methods are close, and most of their mistakes fall in the bins next to $\tau$. For Llama-3.1-8B, ATC does worst, predicting 42--55\% of the datasets in each truly unreliable bin as reliable. Our estimator predicts more of the truly reliable datasets as reliable (61--96\% below the top bin, against 40--70\% for the meta-model).

\begin{table}[t]
\centering
\footnotesize
\setlength{\tabcolsep}{2pt}
\resizebox{\columnwidth}{!}{%
\begin{tabular}{l cc cc cc}
\toprule
 & \multicolumn{2}{c}{GPT-4o mini} & \multicolumn{2}{c}{Llama-3.1-8B} & \multicolumn{2}{c}{Qwen3-8B} \\
\cmidrule(lr){2-3} \cmidrule(lr){4-5} \cmidrule(lr){6-7}
Method & Acc & F1 & Acc & F1 & Acc & F1 \\
\midrule
\textit{All datasets} & \multicolumn{2}{c}{\textit{461 datasets}} & \multicolumn{2}{c}{\textit{461 datasets}} & \multicolumn{2}{c}{\textit{461 datasets}} \\
\textit{Baseline} & \textit{.740} & \textit{.850} & \textit{.638} & \textit{.532} & \textit{.544} & \textit{.705} \\
ATC & $.782_{.005}$ & $.852_{.003}$ & $.539_{.005}$ & $.470_{.005}$ & $.711_{.004}$ & $.718_{.004}$ \\
Meta-model & $.806_{.019}$ & $.866_{.013}$ & $.750_{.014}$ & $.599_{.025}$ & $.705_{.003}$ & $.718_{.005}$ \\
Ours & $\mathbf{.814_{.006}}$ & $\mathbf{.875_{.004}}$ & $\mathbf{.802_{.010}}$ & $\mathbf{.730_{.013}}$ & $\mathbf{.729_{.005}}$ & $\mathbf{.740_{.006}}$ \\
\midrule
\textit{Excluding $R(D) \in [.89, .91)$} & \multicolumn{2}{c}{\textit{419 datasets}} & \multicolumn{2}{c}{\textit{415 datasets}} & \multicolumn{2}{c}{\textit{408 datasets}} \\
\textit{Baseline} & \textit{.754} & \textit{.860} & \textit{.646} & \textit{.523} & \textit{.547} & \textit{.707} \\
ATC & $.814_{.002}$ & $.876_{.002}$ & $.550_{.006}$ & $.482_{.005}$ & $.736_{.007}$ & $.747_{.007}$ \\
Meta-model & $.840_{.019}$ & $.892_{.013}$ & $.761_{.016}$ & $.610_{.027}$ & $.728_{.003}$ & $.744_{.005}$ \\
HGB with $\uncertainty_{\mathrm{first}}$ & $\mathbf{.846_{.006}}$ & $\mathbf{.898_{.004}}$ & $\mathbf{.823_{.009}}$ & $\mathbf{.756_{.012}}$ & $\mathbf{.757_{.006}}$ & $\mathbf{.773_{.005}}$ \\
\midrule
\textit{Excluding $R(D) \in [.875, .925)$} & \multicolumn{2}{c}{\textit{351 datasets}} & \multicolumn{2}{c}{\textit{311 datasets}} & \multicolumn{2}{c}{\textit{309 datasets}} \\
\textit{Baseline} & \textit{.761} & \textit{.864} & \textit{.707} & \textit{.453} & \textit{.518} & \textit{.682} \\
ATC & $.862_{.007}$ & $.910_{.005}$ & $.527_{.005}$ & $.422_{.006}$ & $.811_{.006}$ & $.814_{.006}$ \\
Meta-model & $.871_{.016}$ & $.915_{.011}$ & $.807_{.016}$ & $.649_{.032}$ & $.792_{.009}$ & $.797_{.013}$ \\
HGB with $\uncertainty_{\mathrm{first}}$ & $\mathbf{.889_{.009}}$ & $\mathbf{.928_{.006}}$ & $\mathbf{.851_{.008}}$ & $\mathbf{.769_{.010}}$ & $\mathbf{.830_{.010}}$ & $\mathbf{.835_{.009}}$ \\
\midrule
\textit{Excluding $R(D) \in [.85, .95)$} & \multicolumn{2}{c}{\textit{219 datasets}} & \multicolumn{2}{c}{\textit{196 datasets}} & \multicolumn{2}{c}{\textit{180 datasets}} \\
\textit{Baseline} & \textit{.744} & \textit{.853} & \textit{.821} & \textit{.303} & \textit{.600} & \textit{.571} \\
ATC & $.909_{.008}$ & $.940_{.005}$ & $.517_{.007}$ & $.336_{.008}$ & $.863_{.008}$ & $.836_{.011}$ \\
Meta-model & $.916_{.012}$ & $.945_{.008}$ & $.862_{.009}$ & $.651_{.031}$ & $.835_{.005}$ & $.806_{.008}$ \\
HGB with $\uncertainty_{\mathrm{first}}$ & $\mathbf{.936_{.003}}$ & $\mathbf{.958_{.002}}$ & $\mathbf{.881_{.015}}$ & $\mathbf{.744_{.026}}$ & $\mathbf{.883_{.011}}$ & $\mathbf{.864_{.012}}$ \\
\bottomrule
\end{tabular}}
\caption{Accuracy and F1 at $\tau = 0.9$ and $\kappa = 210$ after excluding the datasets whose true accuracy $R(D)$ lies within a band around $\tau$, with the number of remaining datasets for each LLM. The baseline is recomputed on the remaining datasets and predicts the majority class for accuracy and every dataset as reliable for F1. Values are means over three random seeds with standard errors as subscripts.}
\label{tab:bands}
\end{table}

\begin{table*}[t]
\centering
\footnotesize
\setlength{\tabcolsep}{2pt}
\resizebox{\textwidth}{!}{%
\begin{tabular}{ll cccccc cccccc}
\toprule
 & & \multicolumn{6}{c}{Accuracy} & \multicolumn{6}{c}{F1} \\
\cmidrule(lr){3-8} \cmidrule(lr){9-14}
Classifier $g$ & Signals & 10 & 50 & 90 & 130 & 170 & 210 & 10 & 50 & 90 & 130 & 170 & 210 \\
\splitrule
\multicolumn{2}{l}{\textit{Baseline}} & \multicolumn{6}{c}{\textit{.740}} & \multicolumn{6}{c}{\textit{.850}} \\
\multicolumn{2}{l}{ATC \citep{garg2022leveraging}} & $.717_{.015}$ & $.750_{.016}$ & $.766_{.004}$ & $.774_{.011}$ & $.781_{.012}$ & $.782_{.005}$ & $.815_{.011}$ & $.830_{.012}$ & $.841_{.003}$ & $.847_{.008}$ & $.851_{.008}$ & $.852_{.003}$ \\
\multicolumn{2}{l}{Meta-model \citep{fu-etal-2023-estimating}} & $.666_{.015}$ & $.772_{.016}$ & $\mathbf{.790_{.017}}$ & $.795_{.015}$ & $.810_{.008}$ & $.806_{.019}$ & $.760_{.009}$ & $.846_{.013}$ & $.855_{.013}$ & $.861_{.011}$ & $.871_{.005}$ & $.866_{.013}$ \\
\splitrule
\multirow{6}{*}{HGB} & $\mathrm{uncertainty}_{\mathrm{first}}$ & $.726_{.020}$ & $.767_{.019}$ & $.787_{.005}$ & $\mathbf{.808_{.008}}$ & $.808_{.003}$ & $\mathbf{.814_{.006}}$ & $.822_{.015}$ & $.844_{.015}$ & $.856_{.005}$ & $\mathbf{.872_{.006}}$ & $.871_{.002}$ & $\mathbf{.875_{.004}}$ \\
 & $\mathrm{uncertainty}_{\mathrm{avg}}$ & $.714_{.014}$ & $.762_{.011}$ & $.762_{.011}$ & $.781_{.006}$ & $.787_{.005}$ & $.790_{.002}$ & $.813_{.010}$ & $.841_{.008}$ & $.840_{.008}$ & $.854_{.005}$ & $.857_{.003}$ & $.859_{.000}$ \\
 & agreement & $\mathbf{.754_{.003}}$ & $.769_{.005}$ & $.769_{.002}$ & $.774_{.006}$ & $.773_{.002}$ & $.773_{.001}$ & $\mathbf{.852_{.002}}$ & $\mathbf{.863_{.002}}$ & $\mathbf{.864_{.001}}$ & $.867_{.003}$ & $.866_{.001}$ & $.866_{.000}$ \\
 & confidence & $.677_{.003}$ & $.662_{.008}$ & $.648_{.005}$ & $.659_{.008}$ & $.645_{.008}$ & $.638_{.005}$ & $.786_{.003}$ & $.764_{.008}$ & $.748_{.005}$ & $.760_{.008}$ & $.748_{.006}$ & $.742_{.005}$ \\
 & stability & $.673_{.018}$ & $.661_{.010}$ & $.653_{.008}$ & $.658_{.006}$ & $.657_{.007}$ & $.644_{.005}$ & $.782_{.015}$ & $.762_{.010}$ & $.752_{.009}$ & $.758_{.005}$ & $.756_{.005}$ & $.745_{.005}$ \\
 & All five signals & $.719_{.018}$ & $.757_{.007}$ & $.759_{.008}$ & $.782_{.006}$ & $.779_{.004}$ & $.787_{.005}$ & $.821_{.013}$ & $.838_{.005}$ & $.837_{.006}$ & $.853_{.005}$ & $.851_{.002}$ & $.855_{.003}$ \\
\splitrule
\multirow{6}{*}{XGBoost} & $\mathrm{uncertainty}_{\mathrm{first}}$ & $.726_{.022}$ & $\mathbf{.776_{.013}}$ & $.784_{.008}$ & $.806_{.008}$ & $\mathbf{.811_{.005}}$ & $.810_{.006}$ & $.822_{.016}$ & $.849_{.010}$ & $.854_{.006}$ & $.870_{.006}$ & $\mathbf{.873_{.003}}$ & $.872_{.004}$ \\
 & $\mathrm{uncertainty}_{\mathrm{avg}}$ & $.734_{.026}$ & $.758_{.006}$ & $.761_{.007}$ & $.779_{.008}$ & $.789_{.009}$ & $.786_{.006}$ & $.826_{.019}$ & $.837_{.005}$ & $.839_{.005}$ & $.851_{.006}$ & $.857_{.007}$ & $.855_{.003}$ \\
 & agreement & $\mathbf{.754_{.003}}$ & $.769_{.005}$ & $.769_{.002}$ & $.774_{.006}$ & $.773_{.002}$ & $.773_{.001}$ & $\mathbf{.852_{.002}}$ & $\mathbf{.863_{.002}}$ & $\mathbf{.864_{.001}}$ & $.867_{.003}$ & $.866_{.001}$ & $.866_{.000}$ \\
 & confidence & $.679_{.010}$ & $.662_{.007}$ & $.654_{.010}$ & $.655_{.001}$ & $.647_{.008}$ & $.636_{.006}$ & $.788_{.008}$ & $.764_{.008}$ & $.755_{.010}$ & $.756_{.003}$ & $.750_{.005}$ & $.740_{.005}$ \\
 & stability & $.675_{.013}$ & $.664_{.008}$ & $.660_{.009}$ & $.662_{.006}$ & $.659_{.007}$ & $.650_{.007}$ & $.783_{.011}$ & $.763_{.008}$ & $.759_{.009}$ & $.761_{.004}$ & $.760_{.005}$ & $.751_{.006}$ \\
 & All five signals & $.720_{.008}$ & $.763_{.012}$ & $.775_{.007}$ & $.780_{.007}$ & $.780_{.003}$ & $.785_{.001}$ & $.818_{.008}$ & $.841_{.008}$ & $.847_{.005}$ & $.852_{.005}$ & $.850_{.002}$ & $.854_{.001}$ \\
\splitrule
\multirow{6}{*}{SVM} & $\mathrm{uncertainty}_{\mathrm{first}}$ & $.731_{.016}$ & $.764_{.014}$ & $.787_{.007}$ & $\mathbf{.808_{.008}}$ & $.810_{.008}$ & $.813_{.003}$ & $.824_{.012}$ & $.840_{.010}$ & $.855_{.005}$ & $.870_{.006}$ & $.871_{.005}$ & $.873_{.002}$ \\
 & $\mathrm{uncertainty}_{\mathrm{avg}}$ & $.717_{.015}$ & $.750_{.016}$ & $.766_{.004}$ & $.774_{.011}$ & $.781_{.012}$ & $.782_{.005}$ & $.815_{.011}$ & $.830_{.012}$ & $.841_{.003}$ & $.847_{.008}$ & $.851_{.008}$ & $.852_{.003}$ \\
 & agreement & $\mathbf{.754_{.003}}$ & $.769_{.005}$ & $.769_{.002}$ & $.774_{.006}$ & $.773_{.002}$ & $.773_{.001}$ & $\mathbf{.852_{.002}}$ & $\mathbf{.863_{.002}}$ & $\mathbf{.864_{.001}}$ & $.867_{.003}$ & $.866_{.001}$ & $.866_{.000}$ \\
 & confidence & $.670_{.008}$ & $.655_{.007}$ & $.651_{.008}$ & $.640_{.003}$ & $.646_{.004}$ & $.634_{.008}$ & $.779_{.008}$ & $.759_{.007}$ & $.753_{.007}$ & $.743_{.004}$ & $.748_{.002}$ & $.738_{.006}$ \\
 & stability & $.669_{.009}$ & $.663_{.014}$ & $.661_{.006}$ & $.659_{.009}$ & $.657_{.005}$ & $.650_{.006}$ & $.778_{.007}$ & $.765_{.014}$ & $.761_{.006}$ & $.759_{.007}$ & $.758_{.004}$ & $.751_{.006}$ \\
 & All five signals & $.697_{.016}$ & $.693_{.014}$ & $.701_{.016}$ & $.698_{.003}$ & $.705_{.004}$ & $.698_{.001}$ & $.799_{.012}$ & $.789_{.012}$ & $.791_{.013}$ & $.789_{.003}$ & $.795_{.002}$ & $.789_{.001}$ \\
\bottomrule
\end{tabular}}
\caption{Accuracy and F1 of the estimator for \textbf{GPT-4o mini} at $\tau = 0.9$ with each instance-level classifier $g$, using each signal on its own or all five signals together, for each $\kappa$. Values are means over three random seeds with standard errors as subscripts. }
\label{tab:crossed-gpt}
\end{table*}

\section{Full Results}
\label{app:full}

\tabref{tab:bands} compares our estimator with the prior methods after excluding the datasets closest to $\tau$, it reports metrics at $\kappa = 210$ after excluding the datasets with .01, .025 and .05 of $\tau$, with the baseline recomputed on the remaining datasets.

\begin{figure*}[t]
\centering
\includegraphics[width=0.9\textwidth]{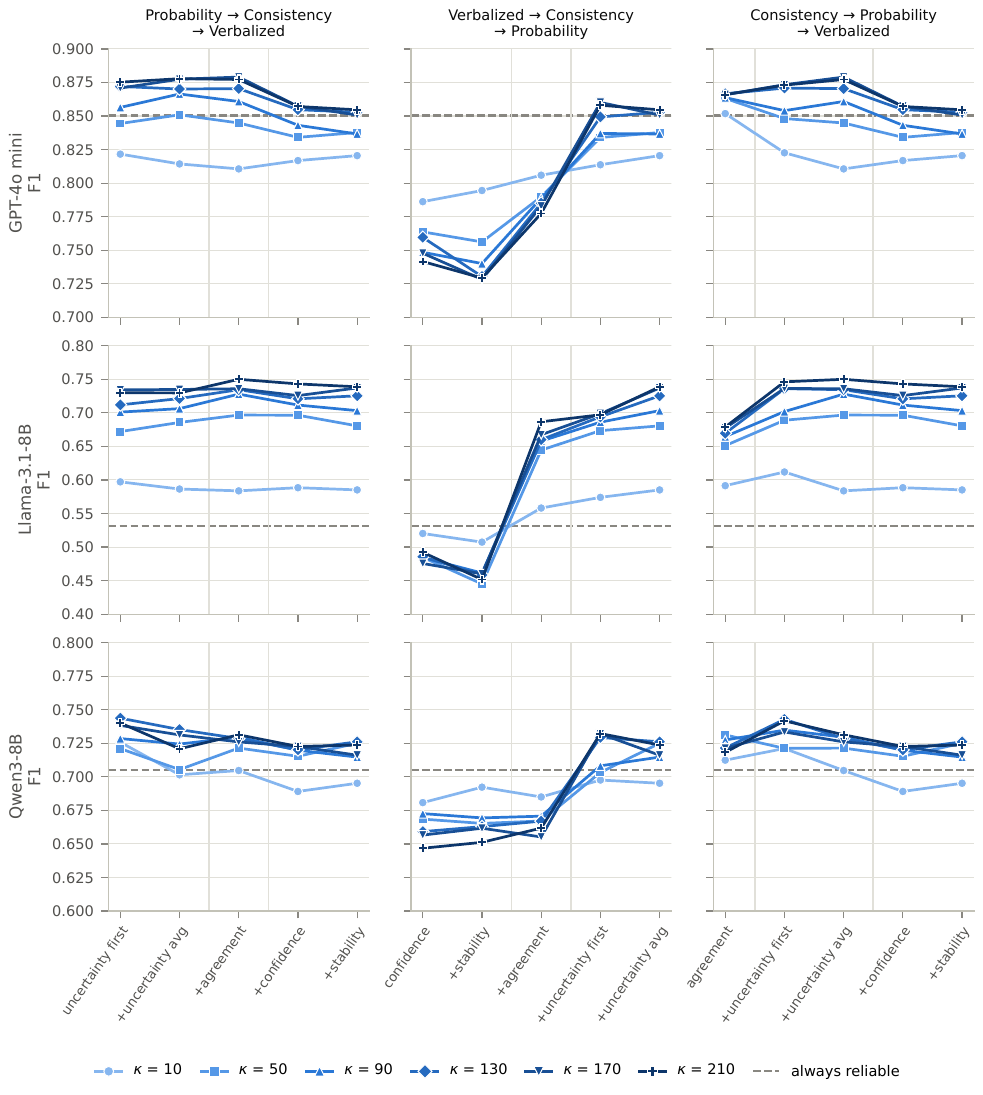}
\caption{F1 of the estimator (HGB) for each LLM at $\tau = 0.9$ as signals are added one at a time (left to right), in three orders that each start from a different signal family; each line is one number of sampled instances $\kappa$. The dashed line is the baseline that predicts every dataset as reliable.}
\label{fig:incremental-all}
\end{figure*}

Tables~\ref{tab:crossed-gpt}--\ref{tab:crossed-qwen} give accuracy and F1 for every combination of instance-level classifier and signal set, one table per LLM. \figref{fig:incremental-all} shows the three signal orders of \figref{fig:incremental}a for all three LLMs.

From $\kappa = 130$ on, the first-token uncertainty is the best single signal for every LLM and classifier, in both accuracy and F1, and with it alone the three classifiers are within .036 of each other. Combining all five signals beats the first-token uncertainty alone at most $\kappa$ only for Llama-3.1-8B. The verbalized signals stay below the F1 baseline at every $\kappa$ for GPT-4o mini and Llama-3.1-8B; for Qwen3-8B, stability is at or slightly above it.

\begin{table*}[t]
\centering
\footnotesize
\setlength{\tabcolsep}{2pt}
\resizebox{\textwidth}{!}{%
\begin{tabular}{ll cccccc cccccc}
\toprule
 & & \multicolumn{6}{c}{Accuracy} & \multicolumn{6}{c}{F1} \\
\cmidrule(lr){3-8} \cmidrule(lr){9-14}
Classifier $g$ & Signals & 10 & 50 & 90 & 130 & 170 & 210 & 10 & 50 & 90 & 130 & 170 & 210 \\
\splitrule
\multicolumn{2}{l}{\textit{Baseline}} & \multicolumn{6}{c}{\textit{.638}} & \multicolumn{6}{c}{\textit{.532}} \\
\multicolumn{2}{l}{ATC \citep{garg2022leveraging}} & $.521_{.015}$ & $.549_{.011}$ & $.547_{.005}$ & $.538_{.005}$ & $.542_{.004}$ & $.539_{.005}$ & $.521_{.012}$ & $.495_{.013}$ & $.484_{.005}$ & $.474_{.005}$ & $.477_{.006}$ & $.470_{.005}$ \\
\multicolumn{2}{l}{Meta-model \citep{fu-etal-2023-estimating}} & $\mathbf{.643_{.005}}$ & $.731_{.003}$ & $.734_{.005}$ & $.751_{.005}$ & $.754_{.011}$ & $.750_{.014}$ & $.496_{.013}$ & $.591_{.012}$ & $.578_{.016}$ & $.608_{.008}$ & $.620_{.019}$ & $.599_{.025}$ \\
\splitrule
\multirow{6}{*}{HGB} & $\mathrm{uncertainty}_{\mathrm{first}}$ & $.607_{.016}$ & $.735_{.013}$ & $.770_{.002}$ & $.785_{.002}$ & $.803_{.002}$ & $.802_{.010}$ & $.597_{.013}$ & $.672_{.015}$ & $.701_{.003}$ & $.712_{.002}$ & $.734_{.003}$ & $.730_{.013}$ \\
 & $\mathrm{uncertainty}_{\mathrm{avg}}$ & $.580_{.019}$ & $.641_{.002}$ & $.648_{.010}$ & $.667_{.010}$ & $.672_{.015}$ & $.651_{.013}$ & $.571_{.021}$ & $.566_{.011}$ & $.566_{.008}$ & $.588_{.016}$ & $.591_{.014}$ & $.564_{.011}$ \\
 & agreement & $.552_{.015}$ & $.645_{.014}$ & $.664_{.005}$ & $.670_{.007}$ & $.680_{.006}$ & $.683_{.004}$ & $.592_{.011}$ & $.651_{.011}$ & $.665_{.006}$ & $.670_{.004}$ & $.677_{.006}$ & $.679_{.005}$ \\
 & confidence & $.502_{.005}$ & $.540_{.015}$ & $.543_{.010}$ & $.548_{.009}$ & $.541_{.007}$ & $.550_{.003}$ & $.520_{.010}$ & $.485_{.011}$ & $.484_{.011}$ & $.486_{.004}$ & $.476_{.008}$ & $.492_{.006}$ \\
 & stability & $.508_{.007}$ & $.573_{.009}$ & $.563_{.001}$ & $.561_{.005}$ & $.563_{.002}$ & $.556_{.001}$ & $.513_{.006}$ & $.507_{.004}$ & $.488_{.008}$ & $.471_{.013}$ & $.465_{.002}$ & $.459_{.008}$ \\
 & All five signals & $.566_{.012}$ & $.750_{.023}$ & $.781_{.006}$ & $\mathbf{.797_{.010}}$ & $.805_{.009}$ & $.810_{.009}$ & $.585_{.008}$ & $.681_{.021}$ & $.703_{.005}$ & $\mathbf{.725_{.011}}$ & $\mathbf{.737_{.011}}$ & $.739_{.012}$ \\
\splitrule
\multirow{6}{*}{XGBoost} & $\mathrm{uncertainty}_{\mathrm{first}}$ & $.603_{.023}$ & $.743_{.015}$ & $.772_{.006}$ & $.786_{.005}$ & $.798_{.013}$ & $.800_{.010}$ & $.597_{.017}$ & $.682_{.017}$ & $.700_{.005}$ & $.715_{.004}$ & $.728_{.014}$ & $.731_{.011}$ \\
 & $\mathrm{uncertainty}_{\mathrm{avg}}$ & $.586_{.021}$ & $.666_{.011}$ & $.672_{.008}$ & $.675_{.004}$ & $.689_{.002}$ & $.676_{.004}$ & $.576_{.015}$ & $.592_{.005}$ & $.577_{.008}$ & $.583_{.008}$ & $.602_{.003}$ & $.581_{.006}$ \\
 & agreement & $.552_{.015}$ & $.644_{.015}$ & $.664_{.005}$ & $.670_{.007}$ & $.680_{.006}$ & $.683_{.004}$ & $.592_{.011}$ & $.651_{.012}$ & $.665_{.006}$ & $.670_{.004}$ & $.677_{.006}$ & $.679_{.005}$ \\
 & confidence & $.495_{.013}$ & $.537_{.008}$ & $.548_{.015}$ & $.548_{.007}$ & $.547_{.010}$ & $.558_{.009}$ & $.516_{.011}$ & $.479_{.010}$ & $.494_{.014}$ & $.487_{.009}$ & $.486_{.010}$ & $.504_{.011}$ \\
 & stability & $.518_{.006}$ & $.568_{.010}$ & $.563_{.003}$ & $.569_{.004}$ & $.562_{.003}$ & $.550_{.006}$ & $.523_{.004}$ & $.498_{.022}$ & $.482_{.006}$ & $.480_{.010}$ & $.465_{.004}$ & $.454_{.004}$ \\
 & All five signals & $.595_{.020}$ & $.753_{.027}$ & $\mathbf{.798_{.011}}$ & $.789_{.004}$ & $\mathbf{.808_{.004}}$ & $\mathbf{.811_{.009}}$ & $.593_{.012}$ & $\mathbf{.689_{.029}}$ & $\mathbf{.725_{.010}}$ & $.715_{.006}$ & $.736_{.005}$ & $\mathbf{.741_{.013}}$ \\
\splitrule
\multirow{6}{*}{SVM} & $\mathrm{uncertainty}_{\mathrm{first}}$ & $.615_{.018}$ & $.732_{.008}$ & $.768_{.007}$ & $.777_{.011}$ & $.784_{.003}$ & $.787_{.008}$ & $.602_{.012}$ & $.653_{.012}$ & $.687_{.008}$ & $.692_{.013}$ & $.698_{.002}$ & $.702_{.007}$ \\
 & $\mathrm{uncertainty}_{\mathrm{avg}}$ & $.521_{.015}$ & $.549_{.011}$ & $.547_{.005}$ & $.538_{.005}$ & $.542_{.004}$ & $.539_{.005}$ & $.521_{.012}$ & $.495_{.013}$ & $.484_{.005}$ & $.474_{.005}$ & $.477_{.006}$ & $.470_{.005}$ \\
 & agreement & $.555_{.015}$ & $.647_{.014}$ & $.665_{.005}$ & $.671_{.006}$ & $.680_{.006}$ & $.684_{.005}$ & $.593_{.011}$ & $.653_{.011}$ & $.665_{.005}$ & $.670_{.004}$ & $.677_{.006}$ & $.679_{.005}$ \\
 & confidence & $.498_{.007}$ & $.513_{.004}$ & $.533_{.012}$ & $.539_{.004}$ & $.530_{.005}$ & $.540_{.009}$ & $.513_{.009}$ & $.467_{.010}$ & $.482_{.016}$ & $.480_{.008}$ & $.466_{.006}$ & $.483_{.011}$ \\
 & stability & $.504_{.007}$ & $.566_{.005}$ & $.570_{.004}$ & $.574_{.012}$ & $.567_{.004}$ & $.558_{.003}$ & $.508_{.003}$ & $.483_{.013}$ & $.480_{.005}$ & $.475_{.014}$ & $.460_{.005}$ & $.448_{.002}$ \\
 & All five signals & $.627_{.019}$ & $\mathbf{.756_{.011}}$ & $.783_{.019}$ & $.795_{.008}$ & $.797_{.006}$ & $.797_{.005}$ & $\mathbf{.609_{.014}}$ & $.674_{.011}$ & $.702_{.022}$ & $.709_{.013}$ & $.711_{.008}$ & $.705_{.007}$ \\
\bottomrule
\end{tabular}}
\caption{Accuracy and F1 of the estimator for \textbf{Llama-3.1-8B} at $\tau = 0.9$ with each instance-level classifier $g$, using each signal on its own or all five signals together, for each $\kappa$. Values are means over three random seeds, with standard errors as subscripts.}
\label{tab:crossed-llama}
\end{table*}

\begin{table*}[t]
\centering
\footnotesize
\setlength{\tabcolsep}{2pt}
\resizebox{\textwidth}{!}{%
\begin{tabular}{ll cccccc cccccc}
\toprule
 & & \multicolumn{6}{c}{Accuracy} & \multicolumn{6}{c}{F1} \\
\cmidrule(lr){3-8} \cmidrule(lr){9-14}
Classifier $g$ & Signals & 10 & 50 & 90 & 130 & 170 & 210 & 10 & 50 & 90 & 130 & 170 & 210 \\
\splitrule
\multicolumn{2}{l}{\textit{Baseline}} & \multicolumn{6}{c}{\textit{.544}} & \multicolumn{6}{c}{\textit{.705}} \\
\multicolumn{2}{l}{ATC \citep{garg2022leveraging}} & $.641_{.020}$ & $.695_{.006}$ & $.698_{.014}$ & $.706_{.004}$ & $.704_{.006}$ & $.711_{.004}$ & $.708_{.013}$ & $.710_{.013}$ & $.710_{.021}$ & $.716_{.002}$ & $.712_{.009}$ & $.718_{.004}$ \\
\multicolumn{2}{l}{Meta-model \citep{fu-etal-2023-estimating}} & $.586_{.022}$ & $.657_{.011}$ & $.683_{.011}$ & $.695_{.010}$ & $.711_{.015}$ & $.705_{.003}$ & $.590_{.034}$ & $.658_{.021}$ & $.684_{.016}$ & $.699_{.010}$ & $.724_{.017}$ & $.718_{.005}$ \\
\splitrule
\multirow{6}{*}{HGB} & $\mathrm{uncertainty}_{\mathrm{first}}$ & $\mathbf{.664_{.027}}$ & $.698_{.005}$ & $.714_{.009}$ & $\mathbf{.729_{.005}}$ & $\mathbf{.726_{.008}}$ & $\mathbf{.729_{.005}}$ & $\mathbf{.726_{.022}}$ & $.721_{.005}$ & $.728_{.015}$ & $\mathbf{.744_{.009}}$ & $\mathbf{.738_{.010}}$ & $\mathbf{.740_{.006}}$ \\
 & $\mathrm{uncertainty}_{\mathrm{avg}}$ & $.627_{.023}$ & $\mathbf{.710_{.008}}$ & $\mathbf{.720_{.007}}$ & $.718_{.006}$ & $.715_{.004}$ & $.713_{.005}$ & $.700_{.022}$ & $.730_{.012}$ & $\mathbf{.735_{.013}}$ & $.731_{.003}$ & $.728_{.003}$ & $.725_{.006}$ \\
 & agreement & $.586_{.014}$ & $.604_{.008}$ & $.595_{.006}$ & $.581_{.005}$ & $.580_{.003}$ & $.573_{.003}$ & $.712_{.007}$ & $\mathbf{.731_{.005}}$ & $.727_{.003}$ & $.721_{.003}$ & $.721_{.002}$ & $.718_{.001}$ \\
 & confidence & $.591_{.015}$ & $.603_{.001}$ & $.612_{.002}$ & $.604_{.005}$ & $.604_{.004}$ & $.597_{.004}$ & $.681_{.004}$ & $.668_{.006}$ & $.673_{.005}$ & $.659_{.002}$ & $.657_{.004}$ & $.647_{.005}$ \\
 & stability & $.589_{.013}$ & $.594_{.009}$ & $.605_{.014}$ & $.613_{.007}$ & $.617_{.009}$ & $.602_{.018}$ & $.698_{.003}$ & $.700_{.003}$ & $.709_{.011}$ & $.713_{.004}$ & $.717_{.005}$ & $.713_{.009}$ \\
 & All five signals & $.618_{.020}$ & $.699_{.010}$ & $.694_{.003}$ & $.706_{.011}$ & $.701_{.004}$ & $.711_{.002}$ & $.695_{.014}$ & $.725_{.006}$ & $.715_{.004}$ & $.726_{.009}$ & $.716_{.005}$ & $.724_{.003}$ \\
\splitrule
\multirow{6}{*}{XGBoost} & $\mathrm{uncertainty}_{\mathrm{first}}$ & $.647_{.035}$ & $.709_{.006}$ & $.713_{.013}$ & $.723_{.004}$ & $.725_{.006}$ & $.727_{.006}$ & $.713_{.024}$ & $.727_{.009}$ & $.725_{.015}$ & $.739_{.005}$ & $.736_{.008}$ & $.738_{.005}$ \\
 & $\mathrm{uncertainty}_{\mathrm{avg}}$ & $.620_{.026}$ & $.698_{.014}$ & $.708_{.003}$ & $.708_{.013}$ & $.708_{.006}$ & $.719_{.006}$ & $.693_{.022}$ & $.718_{.019}$ & $.722_{.007}$ & $.719_{.010}$ & $.719_{.007}$ & $.730_{.008}$ \\
 & agreement & $.587_{.014}$ & $.604_{.008}$ & $.595_{.006}$ & $.581_{.005}$ & $.580_{.003}$ & $.574_{.003}$ & $.713_{.007}$ & $\mathbf{.731_{.005}}$ & $.727_{.003}$ & $.721_{.003}$ & $.721_{.002}$ & $.719_{.002}$ \\
 & confidence & $.593_{.013}$ & $.599_{.003}$ & $.613_{.001}$ & $.607_{.005}$ & $.604_{.003}$ & $.595_{.003}$ & $.682_{.004}$ & $.665_{.008}$ & $.674_{.004}$ & $.662_{.003}$ & $.657_{.003}$ & $.646_{.004}$ \\
 & stability & $.586_{.016}$ & $.601_{.011}$ & $.612_{.010}$ & $.610_{.001}$ & $.618_{.005}$ & $.608_{.021}$ & $.696_{.003}$ & $.706_{.002}$ & $.715_{.008}$ & $.711_{.004}$ & $.716_{.003}$ & $.717_{.011}$ \\
 & All five signals & $.636_{.024}$ & $.684_{.006}$ & $.696_{.003}$ & $.707_{.015}$ & $.708_{.008}$ & $.707_{.003}$ & $.707_{.021}$ & $.710_{.009}$ & $.716_{.008}$ & $.727_{.012}$ & $.724_{.007}$ & $.723_{.004}$ \\
\splitrule
\multirow{6}{*}{SVM} & $\mathrm{uncertainty}_{\mathrm{first}}$ & $.644_{.024}$ & $.707_{.007}$ & $.709_{.008}$ & $.726_{.003}$ & $.725_{.008}$ & $.725_{.006}$ & $.713_{.017}$ & $.725_{.012}$ & $.720_{.011}$ & $.740_{.006}$ & $.735_{.012}$ & $.734_{.006}$ \\
 & $\mathrm{uncertainty}_{\mathrm{avg}}$ & $.641_{.020}$ & $.695_{.006}$ & $.698_{.014}$ & $.706_{.004}$ & $.704_{.006}$ & $.711_{.004}$ & $.708_{.013}$ & $.710_{.013}$ & $.710_{.021}$ & $.716_{.002}$ & $.712_{.009}$ & $.718_{.004}$ \\
 & agreement & $.586_{.014}$ & $.604_{.008}$ & $.595_{.006}$ & $.581_{.005}$ & $.580_{.003}$ & $.573_{.003}$ & $.712_{.007}$ & $\mathbf{.731_{.005}}$ & $.727_{.003}$ & $.721_{.003}$ & $.721_{.002}$ & $.718_{.001}$ \\
 & confidence & $.594_{.015}$ & $.599_{.002}$ & $.610_{.000}$ & $.607_{.003}$ & $.607_{.002}$ & $.597_{.004}$ & $.684_{.007}$ & $.666_{.006}$ & $.672_{.004}$ & $.665_{.004}$ & $.662_{.003}$ & $.651_{.004}$ \\
 & stability & $.553_{.001}$ & $.550_{.003}$ & $.550_{.002}$ & $.550_{.001}$ & $.550_{.001}$ & $.550_{.001}$ & $.705_{.001}$ & $.706_{.001}$ & $.706_{.001}$ & $.707_{.001}$ & $.707_{.001}$ & $.707_{.000}$ \\
 & All five signals & $.623_{.014}$ & $.665_{.012}$ & $.679_{.006}$ & $.674_{.010}$ & $.680_{.006}$ & $.680_{.003}$ & $.691_{.008}$ & $.693_{.009}$ & $.701_{.011}$ & $.696_{.013}$ & $.700_{.009}$ & $.698_{.003}$ \\
\bottomrule
\end{tabular}}
\caption{Accuracy and F1 of the estimator for \textbf{Qwen3-8B} at $\tau = 0.9$ with each instance-level classifier $g$, using each signal on its own or all five signals together, for each $\kappa$. Values are means over three random seeds, with standard errors as subscripts.}
\label{tab:crossed-qwen}
\end{table*}

\end{document}